\documentclass[10pt,twocolumn,letterpaper]{article}
\usepackage{tabularx}
\usepackage{graphicx}
\usepackage{cvpr}
\usepackage[accsupp]{axessibility} % Improves PDF readability for those with disabilities.
\usepackage{times}
\usepackage{epsfig}
\usepackage{graphicx}
\usepackage{amsmath}
\usepackage{amssymb}
\usepackage{hyperref} 
\usepackage{multirow}
\usepackage{booktabs}
\usepackage{xcolor}
\definecolor{darkgreen}{RGB}{0,100,0}

\providecommand{\keywords}[1]
{
  \small	
  \textbf{\textit{Keywords---}} #1
}

\begin{document}

% \tableofcontents

%%%%%%%%% TITLE
\title{AG-VPReID 2025: Aerial-Ground Video-based Person Re-identification Challenge Results}

\author{
\normalsize Kien Nguyen$^{1,*}$, Clinton Fookes$^{1}$, Sridha Sridharan$^{1}$, Huy Nguyen$^{1}$,
\normalsize Feng Liu$^{2}$, Xiaoming Liu$^{3}$, Arun Ross$^{3}$, Dana Michalski$^{4}$,\\
\normalsize Tamás Endrei$^{5,6}$, Ivan DeAndres-Tame$^{5}$, Ruben Tolosana$^{5}$, Ruben Vera-Rodriguez$^{5}$, Aythami Morales$^{5}$, Julian Fierrez$^{5}$, \\
\normalsize Javier Ortega-Garcia$^{5}$, Zijing Gong$^{7}$, Yuhao Wang$^{7}$, Xuehu Liu$^{8}$, Pingping Zhang$^{7}$,
\normalsize Md Rashidunnabi$^{9}$, Hugo Proença$^{9}$, \\
\normalsize Kailash A. Hambarde$^{9}$, Saeid Rezaei$^{10}$\\[3mm]
\normalsize $^{1}$Queensland University of Technology, Australia; \normalsize $^{2}$Drexel University, USA; \normalsize $^{3}$Michigan State University, USA;\\
\normalsize $^{4}$Department of Defence, Australia; \normalsize $^{5}$Universidad Aut\'onoma de Madrid, Spain;\\
\normalsize $^{6}$P\'azm\'any P\'eter Catholic University, Hungary;
\normalsize $^{7}$%School of Future Technology, School of Artificial Intelligence, 
Dalian University of Technology, China; \\
\normalsize $^{8}$%School of Computer Science and Artificial Intelligence, 
Wuhan University of Technology, China;
\normalsize $^{9}$IT: Instituto de Telecomunica\c{c}\~{o}es, 
University of Beira Interior, Portugal; \\
\normalsize $^{10}$University College Cork, Ireland\\[3mm]
\normalsize $^{*}$\texttt{k.nguyenthanh@qut.edu.au} (corresponding author)
}

\maketitle
\thispagestyle{empty}

%%%%%%%%% ABSTRACT
\begin{abstract}
Person re-identification (ReID) across aerial and ground vantage points has become crucial for large-scale surveillance and public safety applications. Although significant progress has been made in ground-only scenarios, bridging the aerial-ground domain gap remains a formidable challenge due to extreme viewpoint differences, scale variations, and occlusions. Building upon the achievements of the AG-ReID 2023 Challenge, this paper introduces the AG-VPReID 2025 Challenge—the first large-scale video-based competition focused on high-altitude (80–120\,m) aerial-ground ReID. Constructed on the new AG-VPReID dataset with 3,027 identities, over 13,500 tracklets, and approximately 3.7 million frames captured from UAVs, CCTV, and wearable cameras, the challenge featured four international teams. These teams developed solutions ranging from multi-stream architectures to transformer-based temporal reasoning and physics-informed modeling. The leading approach, X-TFCLIP from UAM, attained 72.28\% Rank-1 accuracy in the aerial-to-ground ReID setting and 70.77\% in the ground-to-aerial ReID setting, surpassing existing baselines while highlighting the dataset's complexity. For additional details, please refer to the official website at \url{https://agvpreid25.github.io}.
\end{abstract}

%%%%%%%%% BODY TEXT

% \tableofcontents
\section{Introduction}
The problem of automated person re-identification (ReID) involves identifying individuals across multiple non-overlapping cameras \cite{Ye2021DeepLF,mars,GLTR_LS-VID}. It is encountered in surveillance applications, which often have to rely on low-resolution biometric and soft-biometric features to recognize individuals. Person ReID has applications in retail stores, search and rescue operations, healthcare scenarios, and public safety. The recent shift toward {\em video-based} ReID introduces temporal dynamics that enrich identity information by capturing motion and behavior of an individual over time. However, this evolution also presents new challenges, such as variability in motion patterns and maintaining tracking consistency across diverse camera setups.

Airborne platforms and imaging sensors enable aerial-based person ReID, offering advantages in scale, mobility, and observation capabilities \cite{AerialSurveillance}. High-altitude aerial cameras and drone-based cameras capture wider areas with less occlusion than their ground-based counterparts \cite{Li2021UAVHumanAL,AGReIDv1}. They provide operational flexibility for optimal target viewing and can carry multiple sensors (visual, thermal, LiDAR) to improve accuracy and robustness. However, at high altitudes, such as 80–120 meters above ground, challenges become exponentially complex due to extreme scale variations, severe resolution degradation, and dramatic viewpoint differences.

Existing aerial-based person ReID research mainly match aerial images with other aerial images \cite{Kumar2021ThePA, Li2021UAVHumanAL, Zhang2019PersonRI}. Aerial-to-ground (A2G) person re-identification (ReID) remains a relatively underexplored research area, primarily due to its unique challenges such as drastic viewpoint changes, varying poses, and low-resolution imagery. Video-based A2G ReID introduces further complexities, including temporal discontinuities caused by unstable drone tracking, inconsistent motion patterns across viewpoints, and the difficulty of robustly aggregating temporal features under extreme viewing conditions. Further, the absence of large, annotated public video datasets have hindered progress.

To address this gap, we organized the Aerial-to-Ground Video-based Person ReID 2025 (AG-VPReID 2025) challenge. Building on the success of the AG-ReID 2023 Challenge \cite{nguyen2023ag}, which demonstrated community interest and established baseline performance for image-based aerial-ground ReID at 20–60 meters altitude, the new challenge extends to video-based scenarios with challenging operational altitudes. We collected a video dataset comprising over 3.7 million frames and 3,027 identities using cameras from drones at an altitude of 80–120 meters alongside ground-based CCTV and wearable cameras. The dataset captures real-world surveillance with varied imaging conditions, extreme scale variations, and complex temporal dynamics. We annotate 15 soft-biometric attributes per identity, enhancing utility for comprehensive ReID research. The Kaggle-hosted challenge provides live leaderboard tracking for submitted algorithms.

\section{Related Work}
\label{sec:related}
In this section, we provide an overview of existing video-based person ReID datasets and summarize key methodological advances in aerial-ground person ReID.

\subsection{Video-based Person ReID Datasets}
Video-based person ReID datasets have evolved significantly from traditional ground-to-ground configurations to more challenging multi-platform scenarios, as shown in Table~\ref{tab:compare_statics_video}. Early datasets like MARS~\cite{mars}, introduced in 2016, established foundational benchmarks with 1,261 identities, 20,478 tracklets, and 1.19 million frames captured from 6 cameras. Building upon this foundation, LS-VID~\cite{GLTR_LS-VID} expanded the scale substantially with 3,772 identities across 14,943 tracklets and 2.98 million frames from 15 cameras, demonstrating the importance of large-scale data for robust evaluation. Recent efforts have addressed increasingly complex scenarios, including clothing changes and multi-environmental conditions. MEVID~\cite{Davila2023mevid} introduced comprehensive clothing-change scenarios with 158 identities and over 10.46 million frames across 33 cameras collected over 73 days, representing one of the most extensive long-term person ReID datasets, while CCVID~\cite{gu2022CAL} and VCCR~\cite{han20223d} have contributed valuable insights to understanding person ReID under clothing variations with 226 and 392 identities respectively.

The emergence of aerial platforms in video-based person ReID represents a significant paradigm shift from traditional surveillance configurations. P-Destre~\cite{kumar2020p}, representing an early pioneering effort, collected 253 identities from aerial views at relatively low altitudes of 5-6 meters, establishing the foundation for cross-platform ReID research. More recently, G2A-VReID~\cite{zhang2024cross} has advanced the field substantially with 2,788 identities captured from drones operating at moderate altitudes of 20–60 meters alongside ground-based cameras, comprising 5,576 tracklets and 185,907 frames. Our AG-VPReID dataset significantly extends these efforts by featuring unprecedented operational conditions with 3,027 identities, over 13,511 tracklets, and approximately 3.7 million frames captured from multiple platforms including drones operating at extreme altitudes of 80–120 meters, presenting unique challenges due to extreme viewpoint variations, significant scale differences, and complex environmental conditions not addressed in previous datasets.

\begin{table}
{
\fontsize{7}{8}\selectfont
\centering
\caption{AG-VPReID Dataset versus Popular Video-based Person ReID Datasets. ``A/G/W" refers to Aerial, Ground, and Wearable platforms; ``FM" denotes Fixed and Mobile platforms; ``IDs" = Identities, ``Trk." = Tracklets, ``Fr.(M)" = Frames (Million), ``Att." = Attributes, ``Alt." = Altitude, ``CC" = Clothing Change.}
\label{tab:compare_statics_video}
\setlength{\tabcolsep}{2.5pt}
\begin{tabularx}{\columnwidth}{l|c|c|c|c|c|c|c|c}
\toprule
\textbf{Dataset}  & \textbf{Type} & \textbf{IDs} & \textbf{Trk.} & \textbf{Fr.(M)} & \textbf{Att.} & \textbf{Alt.} & \textbf{View} & \textbf{CC} \\
\hline
MARS\cite{mars}               & CCTV     &1,261 &20,478  &1.19 &$\times$         &$<10m$       &Fix & $\times$ \\

LS-VID\cite{GLTR_LS-VID}      & CCTV     &3,772 &14,943  &2.98 &$\times$         &$<10m$       &Fix & $\times$ \\

VCCR\cite{han20223d}          & CCTV     &392   &4,384   &0.15 &$\times$         &$<10m$       &Fix & $\checkmark$ \\

CCVID\cite{gu2022CAL}         & CCTV     &226   &2,856   &0.34 &$\times$         &$<10m$       &Fix & $\checkmark$ \\

MEVID\cite{Davila2023mevid}   & CCTV     &158   &8,092   &10.46 &$\times$        &$<10m$       &Fix & $\checkmark$ \\

P-Destre\cite{kumar2020p}     & UAV      &253   &1,894   &0.10 &7     &5-6m    &Mob & $\times$ \\

G2A-VReID\cite{zhang2024cross} & A/G    &2,788 &5,576   &0.18 &$\times$         &20–60m  &FM & $\times$ \\

\hline
\hline
AG-VPReID  & A/G/W  &3,027 &13,511 &3.7 &15 &80–120m    &FM & $\checkmark$ \\
\bottomrule
\end{tabularx}}
\end{table}

\subsection{Aerial-Ground Person ReID Methods}
Early aerial-ground person ReID research focused on image-based approaches with limited temporal exploration. The AG-ReID 2023 challenge~\cite{nguyen2023ag} established effective cross-platform matching strategies for image-based ReID at 20–60 meter altitudes, with LENS-AG-Net achieving 97.73\% Rank-1 accuracy through HRNet-18, re-ranking, data augmentation, and pseudo-labeling, while CentroidNet attained 94.28\% using centroid representations and BENTO achieved 92.95\% through ensemble techniques. Video-based ReID methods have incorporated temporal information via temporal complementary learning~\cite{hou2020temporal}, Transformer architectures~\cite{liu2021video,he2021dense}, and attention mechanisms~\cite{wang2021pyramid}, though primarily for ground-based scenarios. High-altitude aerial-ground video ReID introduces substantial complexities: extreme scale variations between 80–120 meter aerial and ground footage, temporal discontinuities from unstable drone tracking, motion inconsistencies across viewpoints, and severe resolution degradation limiting feature extraction. While recent work in aerial surveillance~\cite{AerialSurveillance} and multi-modal ReID~\cite{liu2021watching} provides foundational techniques, the combination of high-altitude video with ground footage remains unexplored, with AG-VPReID 2025 addressing this gap through comprehensive benchmarking of algorithms for extreme operational conditions.

\begin{figure*}
    \centering
    \includegraphics[width=0.95\linewidth]{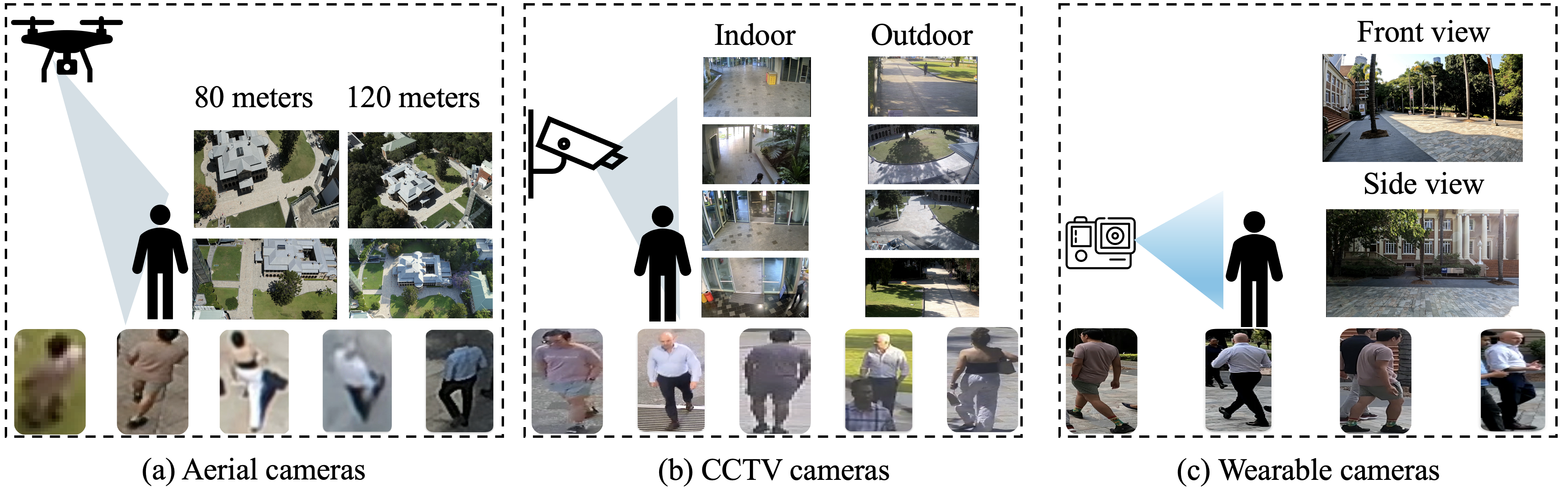}
    \caption{Viewpoint Challenge in our AG-VPReID dataset: extreme elevated view of high-altitude aerial cameras (80–120m), horizontal view of the CCTV camera, and first-person view of the wearable camera, demonstrating the unprecedented scale and viewpoint variations in video-based aerial-ground person re-identification.}
    \label{fig:perspective}
\end{figure*}

\begin{figure}
    \centering
    \includegraphics[width=1\linewidth]{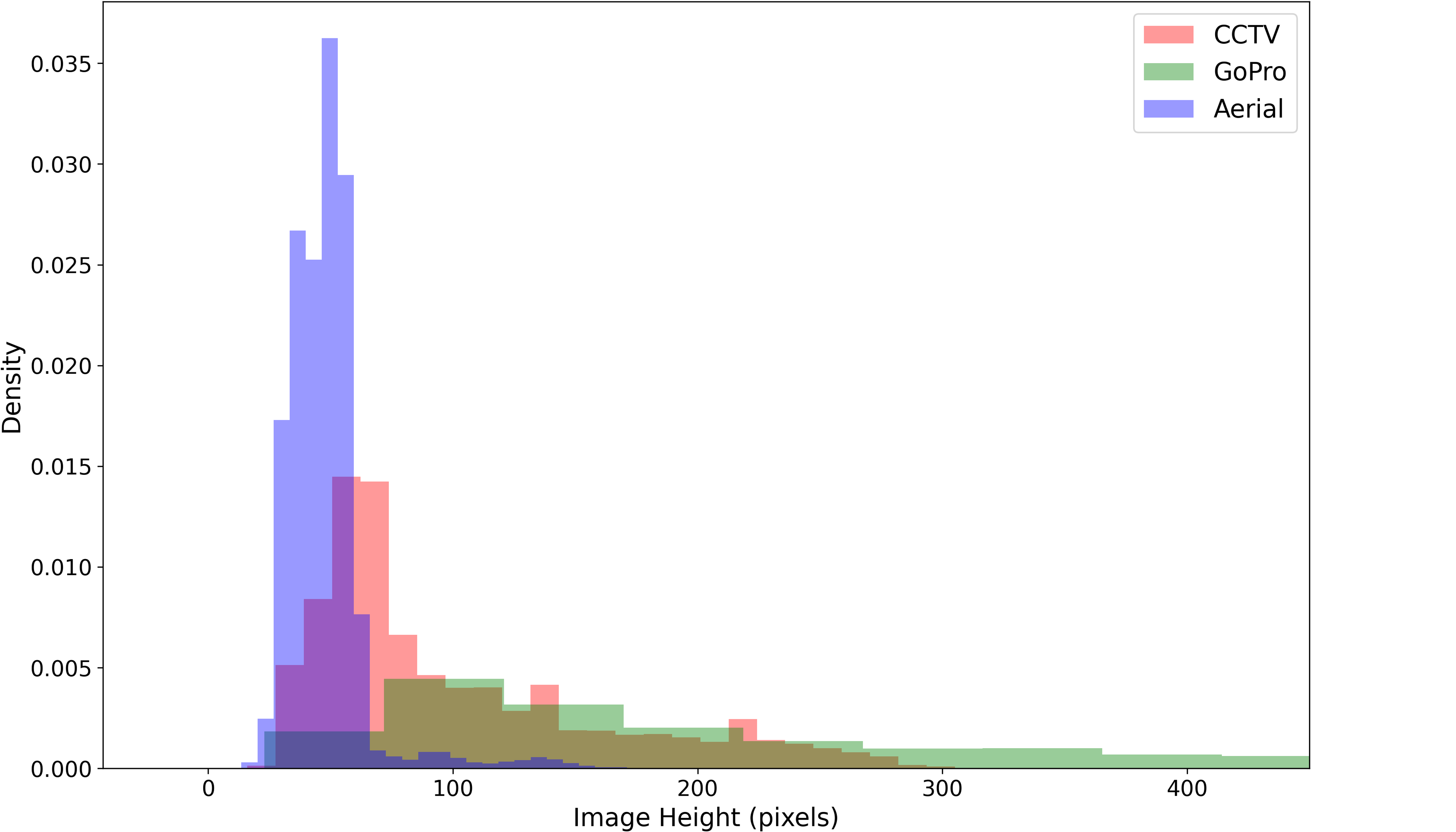}
    \caption{Distributions of the human body heights across three cameras (Aerial, CCTV, GoPro) in the AG-VPReID dataset.}
    \label{fig:height_width_distribution}
\end{figure}

\begin{figure}
    \centering
    \includegraphics[width=0.75\linewidth]{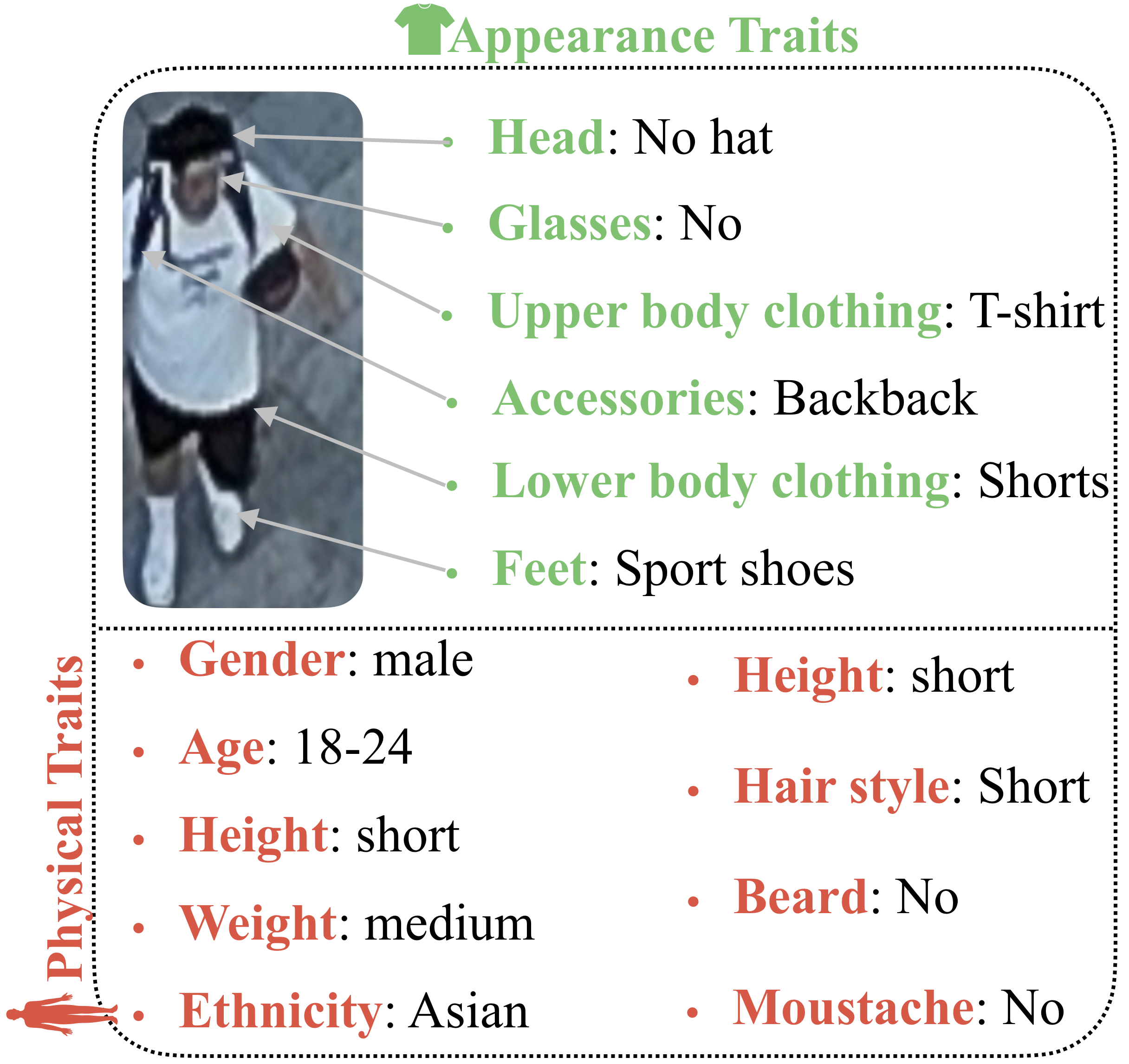}
    \caption{Fifteen soft-biometric labels considered in the AG-VPReID dataset.}
    \label{fig:soft_bio}
\end{figure}

\section{The AG-VPReID2025 Challenge}
The AG-VPReID2025 competition establishes an advanced framework for evaluating video-based person ReID in aerial-ground surveillance environments, extending the AG-ReID2023 image-based competition \cite{nguyen2023ag} with UAVs operating at higher altitudes of 80–120 meters \cite{nguyen2025ag}. This challenge introduces significant technical challenges through extreme viewing disparities, scale variations, temporal discontinuities, and intricate motion patterns requiring innovative algorithmic approaches.

\subsection{Dataset}
\label{sec:dataset}
This challenge utilizes a high-altitude subset of our AG-VPReID dataset \cite{nguyen2025ag}, comprising 3,027 identities with 13,511 tracklets and approximately 3.7 million frames. The dataset employs DJI UAVs (4K-8K resolution) at 80m and 120m, Bosch CCTV cameras (704×480 to 1280×720 pixels), and GoPro Hero10 cameras (4K/1080p resolution). Human body sizes vary dramatically from 18×37 to 293×542 pixels for UAV imagery, posing unprecedented challenges, as illustrated in Figures \ref{fig:perspective} and \ref{fig:height_width_distribution}.

\begin{table}
\fontsize{8}{10}\selectfont
\centering
\caption{Statistics of the testing set for our AG-VPReID dataset.}
\label{tab:static_test_set_agvpreid}
\begin{tabular}{l|c|c|c}
\toprule
\textbf{Subset} & \textbf{Altitude} & \textbf{IDs} & \textbf{Tracklets} \\ 
\hline
\multicolumn{4}{c}{\textit{Testing Case 1: Aerial to Ground (A2G)}} \\
\hline
Query & 80m & 506 & 1,400 \\
Query & 120m & 356 & 1,379 \\
Query & 80m + 120m& 645 & 3,023 \\
\hline
Gallery & Ground & 645 & 2,750 \\
\hline
\hline
\multicolumn{4}{c}{\textit{Testing Case 2: Ground to Aerial (G2A)}} \\
\hline
Query & Ground & 645 & 2,750 \\
\hline
Gallery & 80m & 377 & 1,048 \\
Gallery & 120m & 308 & 1,664 \\
Gallery & 80m + 120m& 645 & 3,023 \\
Gallery Distractor & 80m + 120m& 1,693 & 2,417 \\
\bottomrule
\end{tabular}
\end{table}

% \begin{figure}
%     \centering
%     \includegraphics[width=1\linewidth]{figs/tracklets_distribution.png}
%     \caption{Tracklets distribution across cameras and altitudes.}
%     \label{fig:tracklets_distribution}
% \end{figure}

The dataset includes 15 soft-biometric labels such as gender, age, clothing, and accessories \cite{Nguyen2024AGReIDv2BA}, as shown in Figure \ref{fig:soft_bio}. The dataset is divided into 689 training identities (5,317 tracklets, 1.47M frames) and 645 testing identities across aerial-to-ground and ground-to-aerial scenarios, with 1,693 distractor identities. Detailed testing subset information is provided in Table \ref{tab:static_test_set_agvpreid}.

\begin{figure}
    \centering
    \includegraphics[width=\linewidth]{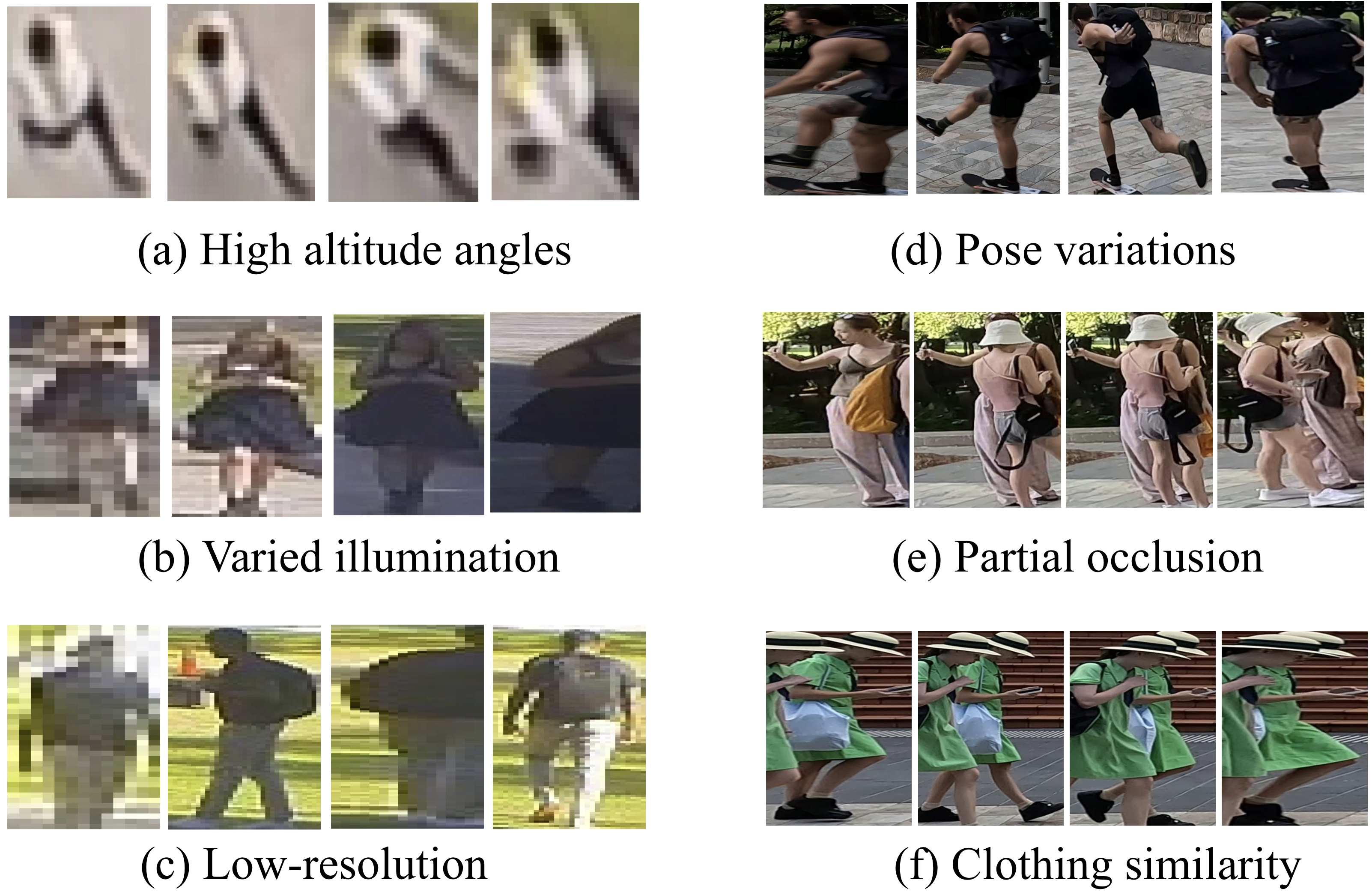}
    \caption{Six key challenges in the AG-VPReID dataset.}
    \label{fig:Challenges}
\end{figure}

Our AG-VPReID2025 dataset poses significant challenges in high-altitude aerial-ground video-based person ReID as follows:
\begin{itemize}
    \item \textit{Extreme Altitude Operations:} UAV operations at 80–120 meters introduce severe scale variations and resolution degradation.
    \item \textit{Temporal Complexity:} Video sequences introduce temporal discontinuities and motion inconsistencies across viewpoints.
    \item \textit{Multi-Platform Integration:} The combination of aerial, ground-based CCTV, and wearable cameras creates diverse resolution and perspective combinations, with extreme variations in subject scale and viewing angles across the three platforms.
    \item \textit{Scale Diversity:} Due to extreme altitude variations and different camera specifications, the sizes of human subjects vary dramatically across sequences, ranging from barely visible at 120m altitude to detailed captures from wearable cameras.
    \item \textit{Environmental Challenges:} Six critical factors—extreme viewpoints, low resolution, occlusions, illumination changes, clothing similarities, and pose variations, as presented in Figure \ref{fig:Challenges}, complicate video-based person ReID.
\end{itemize}

\noindent \textbf{Ethics Approval:} The research has been conducted in accordance with ethical guidelines and has received approval from the relevant institutional review board for data collection and usage of human subject data for research purposes. 

\subsection{Submission}
\label{sec:methods}
The AG-VPReID 2025 Challenge received four algorithmic submissions from international research groups. Each team was permitted a maximum of ten submissions per day on the Kaggle platform with live leaderboard updates. Detailed descriptions are provided in Appendices~\ref{app:X-TFCLIP}, \ref{app:DUT_IIAU_LAB}, \ref{app:ACP-VPReID}, and \ref{app:MFHO}. All teams confirmed adherence to challenge guidelines, using exclusively the official training set without external datasets. X-TFCLIP (Appendix~\ref{app:X-TFCLIP}) emerged as the leading solution by extending TF-CLIP with attention-based feature selection, architectural modifications, and comprehensive loss functions incorporating Online Label Smoothing and Image-to-Text Loss. TFCLIP-AG-VPReID (Appendix~\ref{app:DUT_IIAU_LAB}) secured competitive performance through robust temporal modeling, replacing the original ViT-B-16 image encoder with an EVA-CLIP-L-14 encoder while expanding the training dataset by incorporating the G2AVReID dataset. ACP-VPReID (Appendix~\ref{app:ACP-VPReID}) introduced physics-informed appearance modeling with view-specific encoding, temporal memory diffusion, and motion-aware gating for extreme viewpoint gaps. MFHO (Appendix~\ref{app:MFHO}) employed multi-fidelity hyperparameter optimization to explore the complex parameter space. These approaches demonstrate significant advancements in extreme altitude aerial-ground person ReID, utilizing temporal attention, multi-scale feature extraction, and cross-modal learning to handle unprecedented scale variations and viewpoint differences.

\begin{table*}
\caption{AG-VPReID2025 Challenge Results - Overall Performance. \textcolor{darkgreen}{Green} values indicate the best performance, \textcolor{blue}{blue} values indicate the second-best performance, and \textcolor{red}{red} values indicate the third-best performance in each metric.}
\label{tab:reid-results-overall}
\centering
\fontsize{6}{10}\selectfont
\begin{tabular}{c||c|c|c|c||c|c|c|c||c|c|c|c}
\toprule
\multirow{2}{*}{\textbf{Method}} & \multicolumn{4}{c||}{\textbf{Aerial$\rightarrow$Ground}} & \multicolumn{4}{c||}{\textbf{Ground$\rightarrow$Aerial}} & \multicolumn{4}{c}{\textbf{Overall}}\\
\cline{2-13}
& R1 & R5 & R10 & mAP & R1 & R5 & R10 & mAP & R1 & R5 & R10 & mAP\\
\hline
X-TFCLIP & $\textcolor{darkgreen}{\textbf{72.28}}$ & $\textcolor{darkgreen}{\textbf{81.94}}$ & $\textcolor{darkgreen}{\textbf{85.81}}$ & $\textcolor{darkgreen}{\textbf{74.45}}$ & $\textcolor{blue}{70.77}$ & $\textcolor{blue}{82.59}$ & $\textcolor{blue}{86.08}$ & $\textcolor{blue}{72.67}$& $\textcolor{blue}{71.56}$ & $\textcolor{darkgreen}{\textbf{82.25}}$ & $\textcolor{darkgreen}{\textbf{85.94}}$ & $\textcolor{darkgreen}{\textbf{73.60}}$ \\
TFCLIP-AG-VPReID & $\textcolor{blue}{70.63}$ & $\textcolor{blue}{79.92}$ & $\textcolor{blue}{83.39}$ & $\textcolor{blue}{72.44}$ & $\textcolor{darkgreen}{\textbf{73.28}}$ & $\textcolor{darkgreen}{\textbf{84.08}}$ & $\textcolor{darkgreen}{\textbf{87.20}}$ & $\textcolor{darkgreen}{\textbf{72.99}}$ & $\textcolor{darkgreen}{\textbf{71.89}}$ & $\textcolor{blue}{81.90}$ & $\textcolor{blue}{85.21}$ & $\textcolor{blue}{72.70}$ \\
ACP-VPReID & $\textcolor{red}{65.83}$ & $\textcolor{red}{76.45}$ & $\textcolor{red}{81.34}$ & $\textcolor{red}{68.08}$ & $\textcolor{red}{69.79}$ & $\textcolor{red}{81.64}$ & $\textcolor{red}{85.39}$ & $\textcolor{red}{70.73}$& $\textcolor{red}{67.72}$ & $\textcolor{red}{78.92}$ & $\textcolor{red}{83.27}$ & $\textcolor{red}{69.34}$ \\
MFHO & $64.41$ & $75.09$ & $79.99$ & $66.86$ & $65.50$ & $79.68$ & $83.75$ & $67.83$ & $64.93$ & $77.28$ & $81.78$ & $67.32$ \\
\hline\hline
TF-CLIP & $63.08$ & $75.16$ & $79.89$ & $65.52$ & $64.49$ & $79.86$ & $83.97$ & $67.07$ & $63.75$ & $77.40$ & $81.83$ & $66.26$ \\
\bottomrule
\end{tabular}
\end{table*}

\begin{table}
\caption{AG-VPReID2025 Results - 80m Altitude.}
\label{tab:reid-results-80m}
\centering
\fontsize{6}{10}\selectfont
\begin{tabular}{c||c|c||c|c||c|c}
\toprule
\multirow{2}{*}{\textbf{Method}} & \multicolumn{2}{c||}{\textbf{Aerial$\rightarrow$Ground}} & \multicolumn{2}{c||}{\textbf{Ground$\rightarrow$Aerial}} & \multicolumn{2}{c}{\textbf{Overall}}\\
\cline{2-7}
& R1 & mAP & R1 & mAP & R1 & mAP\\
\hline
TFCLIP-AG-VPReID & $\textcolor{darkgreen}{\textbf{79.67}}$ & $\textcolor{blue}{80.90}$ & $\textcolor{darkgreen}{\textbf{76.86}}$ & $\textcolor{darkgreen}{\textbf{79.09}}$ & $\textcolor{darkgreen}{\textbf{78.47}}$ & $\textcolor{darkgreen}{\textbf{80.13}}$ \\
X-TFCLIP & $\textcolor{blue}{79.29}$ & $\textcolor{darkgreen}{\textbf{81.43}}$ & $\textcolor{blue}{74.59}$ & $\textcolor{blue}{76.97}$ & $\textcolor{blue}{77.28}$ & $\textcolor{blue}{79.52}$ \\
ACP-VPReID & $\textcolor{red}{73.55}$ & $\textcolor{red}{75.98}$ & $\textcolor{red}{73.85}$ & $\textcolor{red}{75.40}$ & $\textcolor{red}{73.68}$ & $\textcolor{red}{75.73}$ \\
MFHO & $74.00$ & $76.22$ & $71.22$ & $73.05$ & $72.81$ & $74.86$ \\
\hline\hline
TF-CLIP & $72.79$ & $75.30$ & $66.98$ & $70.57$ & $70.30$ & $73.27$ \\
\bottomrule
\end{tabular}
\end{table}

\subsection{Evaluation}
\label{sec:protocol}
\begin{figure}
    \centering
    \includegraphics[width=1\linewidth]{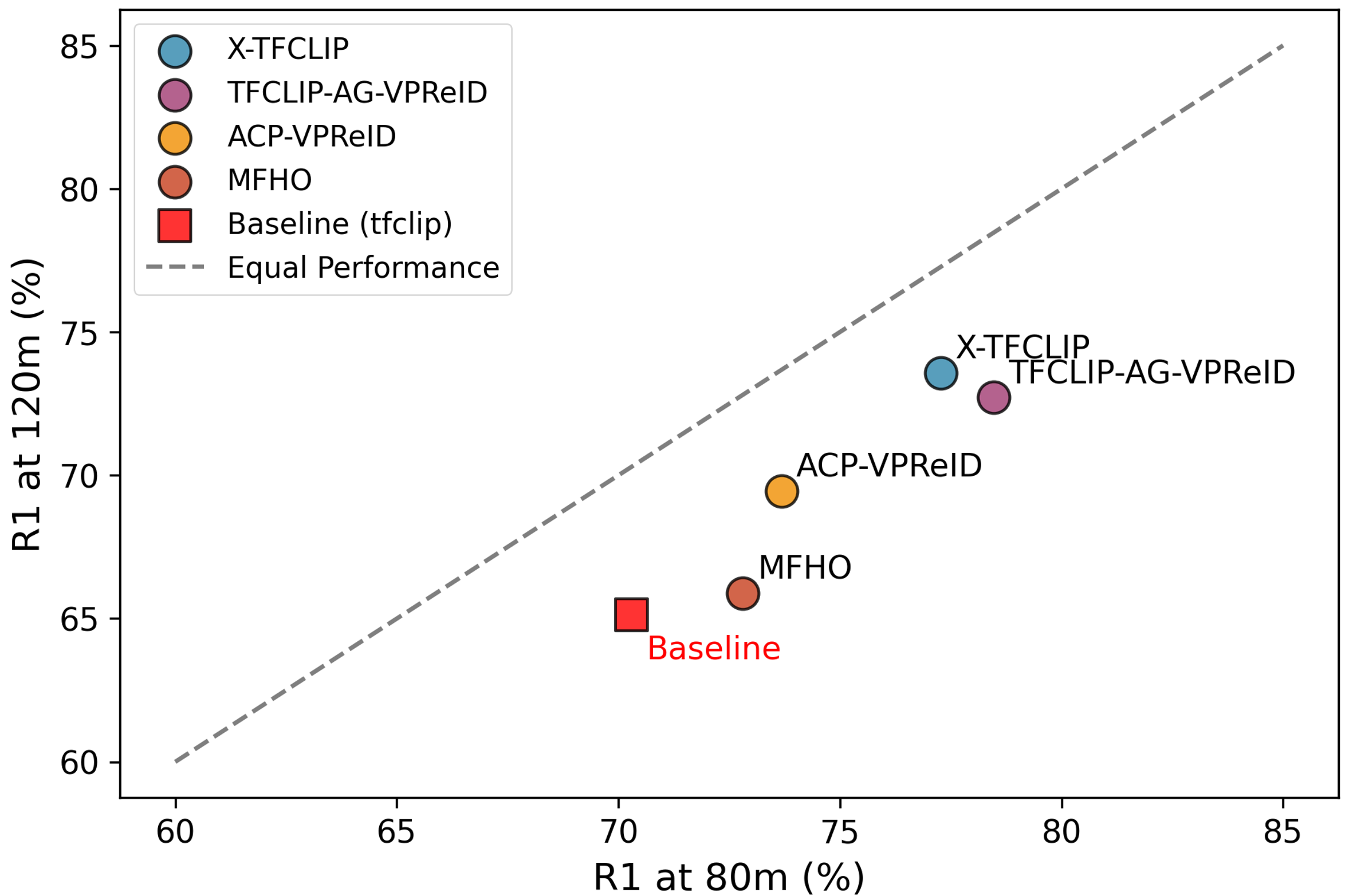}
    \caption{Rank-1 Performance comparison across altitudes.}
    \label{fig:80vs120}
\end{figure}

The summarized results of the AG-VPReID2025 Challenge are compiled in Tables~\ref{tab:reid-results-overall}, \ref{tab:reid-results-80m}, and \ref{tab:reid-results-120m}, presenting the Rank-1 scores, Rank-5 scores, Rank-10 scores, and mean Average Precision (mAP) for each competing method across different altitude scenarios. Notably, all four methods outperform the baseline TF-CLIP approach, indicating significant advancements in the field of high-altitude aerial-ground video-based person ReID. In a competitive challenge, the winning team is X-TFCLIP~(Appendix \ref{app:X-TFCLIP}) achieving the highest overall performance with Rank-1 accuracy of 71.56\% and mAP of 73.60\%, demonstrating the effectiveness of their approach, which combines attention-based feature selection with architectural modifications to the CLIP backbone. TFCLIP-AG-VPReID~(Appendix \ref{app:DUT_IIAU_LAB}) secured a commendable overall Rank-1 accuracy of 71.89\% and mAP of 72.70\%, showcasing particularly strong performance in Ground-to-Aerial scenarios with 73.28\% Rank-1. ACP-VPReID~(Appendix \ref{app:ACP-VPReID}) achieved an overall Rank-1 of 67.72\% and mAP of 69.34\% through their innovative adaptive cross-perspective approach, while MFHO~(Appendix \ref{app:MFHO}) attained 64.93\% overall Rank-1 and 67.32\% mAP via multi-fidelity hyperparameter optimization. Altitude-specific analysis reveals critical performance dependencies, with TFCLIP-AG-VPReID excelling at 80m altitude (78.47\% overall Rank-1 accuracy) while X-TFCLIP maintained superior performance at the more challenging 120m altitude (73.58\% overall Rank-1 accuracy), as shown in Figure~\ref{fig:80vs120}. Most methods experienced 4-6\% Rank-1 accuracy reduction when transitioning from 80m to 120m operations, as shown in Figure~\ref{fig:drop}, highlighting the severe challenges posed by extreme aerial viewpoints. The Ground-to-Aerial matching scenarios generally outperformed Aerial-to-Ground scenarios across all methods, suggesting that upward viewpoint adaptation may be more tractable than downward perspective matching. The consistent improvements demonstrated in the AG-VPReID2025 Challenge signify remarkable progress in video-based person ReID within high-altitude aerial-ground contexts, while the absolute performance levels achieved underscore the dataset's challenging nature and the continued need for algorithmic innovation to address the complex combination of extreme viewpoint variations, severe scale differences, temporal discontinuities, and resolution degradation inherent in high-altitude aerial-ground surveillance scenarios.

\begin{table}
\caption{AG-VPReID2025 Results - 120m Altitude.}
\label{tab:reid-results-120m}
\centering
\fontsize{6}{10}\selectfont
\begin{tabular}{c||c|c||c|c||c|c}
\toprule
\multirow{2}{*}{\textbf{Method}} & \multicolumn{2}{c||}{\textbf{Aerial$\rightarrow$Ground}} & \multicolumn{2}{c||}{\textbf{Ground$\rightarrow$Aerial}} & \multicolumn{2}{c}{\textbf{Overall}}\\
\cline{2-7}
& R1 & mAP & R1 & mAP & R1 & mAP\\
\hline
X-TFCLIP & $\textcolor{darkgreen}{\textbf{73.29}}$ & $\textcolor{darkgreen}{\textbf{75.69}}$ & $\textcolor{blue}{73.82}$ & $\textcolor{darkgreen}{\textbf{76.47}}$ & $\textcolor{darkgreen}{\textbf{73.58}}$ & $\textcolor{darkgreen}{\textbf{76.11}}$ \\
TFCLIP-AG-VPReID & $\textcolor{blue}{69.81}$ & $\textcolor{blue}{72.10}$ & $\textcolor{darkgreen}{\textbf{75.14}}$ & $\textcolor{blue}{75.52}$ & $\textcolor{blue}{72.72}$ & $\textcolor{blue}{73.96}$ \\
ACP-VPReID & $\textcolor{red}{66.01}$ & $\textcolor{red}{69.05}$ & $\textcolor{red}{72.30}$ & $\textcolor{red}{73.77}$ & $\textcolor{red}{69.45}$ & $\textcolor{red}{71.63}$ \\
MFHO & $63.69$ & $66.90$ & $67.71$ & $70.83$ & $65.89$ & $69.05$ \\
\hline\hline
TF-CLIP & $61.71$ & $65.48$ & $67.97$ & $71.26$ & $65.13$ & $68.64$ \\
\bottomrule
\end{tabular}
\end{table}

\section{Discussion}
\label{sec:results}

\begin{figure}
    \centering
    \includegraphics[width=1\linewidth]{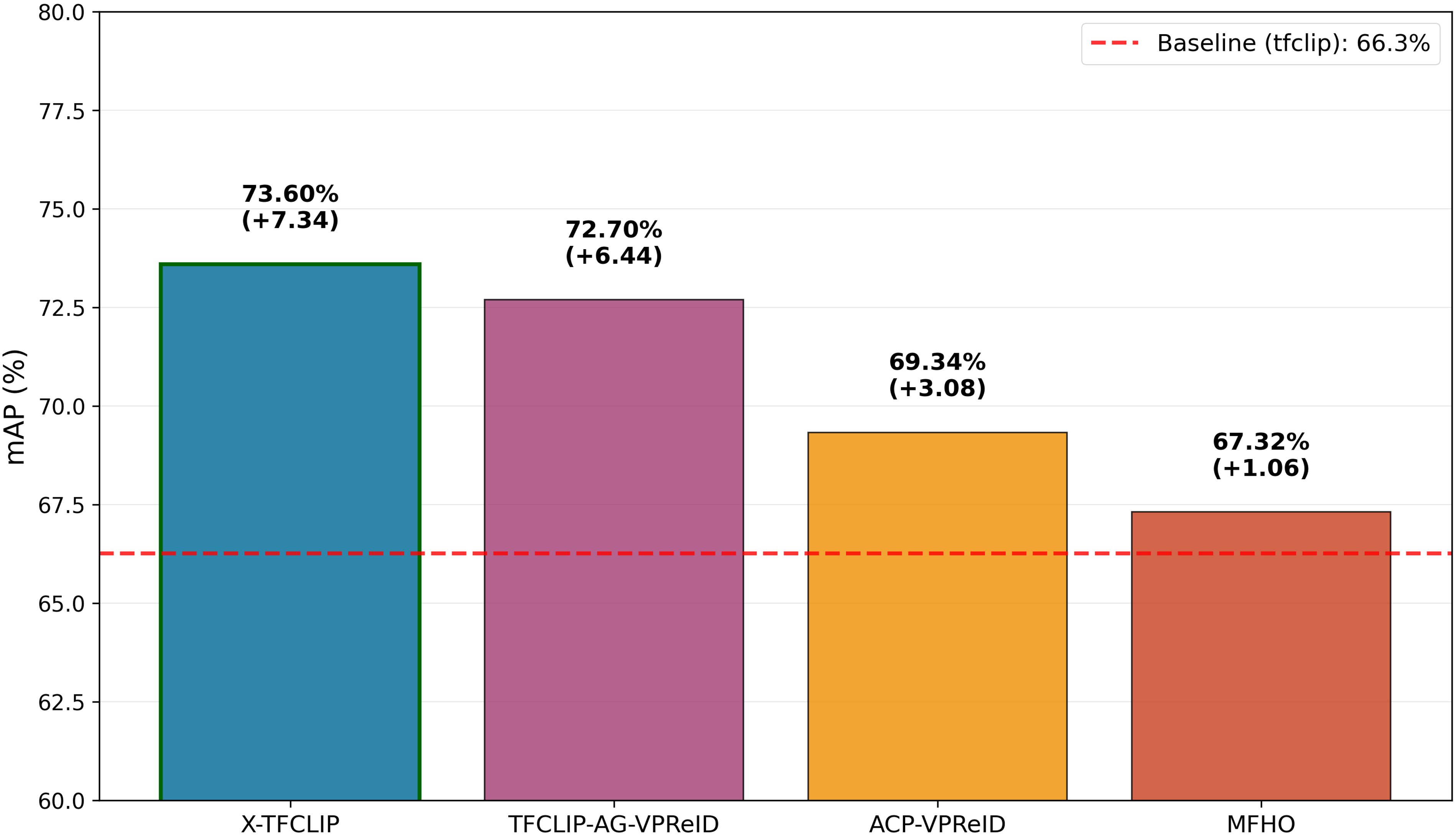}
    \caption{mAP accuracy comparison (Overall).}
    \label{fig:r1_overall}
\end{figure}

From the AG-VPReID2025 Challenge results, as shown in Tables \ref{tab:reid-results-overall}, \ref{tab:reid-results-80m}, and \ref{tab:reid-results-120m}, and Figure~\ref{fig:r1_overall}, the evaluated methods address high-altitude video-based
person ReID challenges with varying effectiveness. These challenges stem from extreme elevated viewpoints, resolution degradation, temporal discontinuities, and dramatic scale variations inherent in high-altitude aerial-ground applications.

\textbf{Extreme Altitude Operations and Method Performance.} X-TFCLIP (Appendix \ref{app:X-TFCLIP}) demonstrated the highest overall performance with 71.56\% Rank-1 accuracy, outperforming other methods particularly in the challenging Aerial-to-Ground scenarios with 72.28\% Rank-1. TFCLIP-AG-VPReID (Appendix \ref{app:DUT_IIAU_LAB}) achieved competitive performance with 71.89\% overall Rank-1, particularly excelling in Ground-to-Aerial scenarios with 73.28\% Rank-1 through its adoption of the stronger EVA-CLIP-L-14 backbone and expanded training dataset incorporating G2AVReID data. This suggests effective handling of unprecedented challenges through attention-based feature selection and robust temporal modeling.

\begin{table*}
\caption{Summary of Methodologies Used in AG-VPReID2025 Challenge.}
\label{tab:methodologies}
\centering
\fontsize{6}{7}\selectfont
\begin{tabularx}{\textwidth}{*1{>{\centering\arraybackslash}X}@{}|| *7{>{\centering\arraybackslash}X} @{}}
\toprule
\textbf{Method} & \textbf{Backbone} & \textbf{Loss Functions} & \textbf{Key Features} & \textbf{Rerank} & \textbf{Ensemble} & \textbf{Cam. Aware} & \textbf{Attr. Aware}\\
\hline
X-TFCLIP (\ref{app:X-TFCLIP}) & CLIP-ViT-B/16 & OLS Cross-entropy, Triplet, I2T & Attention pooling, bicubic extrapolation & $\times$ & $\times$ & \checkmark & \checkmark \\
\hline
TFCLIP-AG-VPReID (\ref{app:DUT_IIAU_LAB}) & EVA-CLIP-L-14 & V2M Contrastive, Triplet, Label Smoothing CE & Stronger backbone, expanded dataset & $\times$ & $\times$ & $\times$ & $\times$ \\
\hline
ACP-VPReID (\ref{app:ACP-VPReID}) & CLIP-ViT-B/16 & Identity, Triplet, Center, Contrastive & View-specific encoding, temporal memory diffusion & $\times$ & $\times$ & \checkmark & $\times$ \\
\hline
MFHO (\ref{app:MFHO}) & CLIP-ViT-B/16 & Hyperband-optimized & Multi-fidelity hyperparameter optimization & - & - & - & - \\
\hline\hline
TF-CLIP (Baseline) & CLIP-ViT-B/16 & Cross-entropy, Triplet & Standard temporal feature aggregation & $\times$ & $\times$ & $\times$ & $\times$ \\
\bottomrule
\end{tabularx}
\end{table*}

\begin{figure}
    \centering
    \includegraphics[width=1\linewidth]{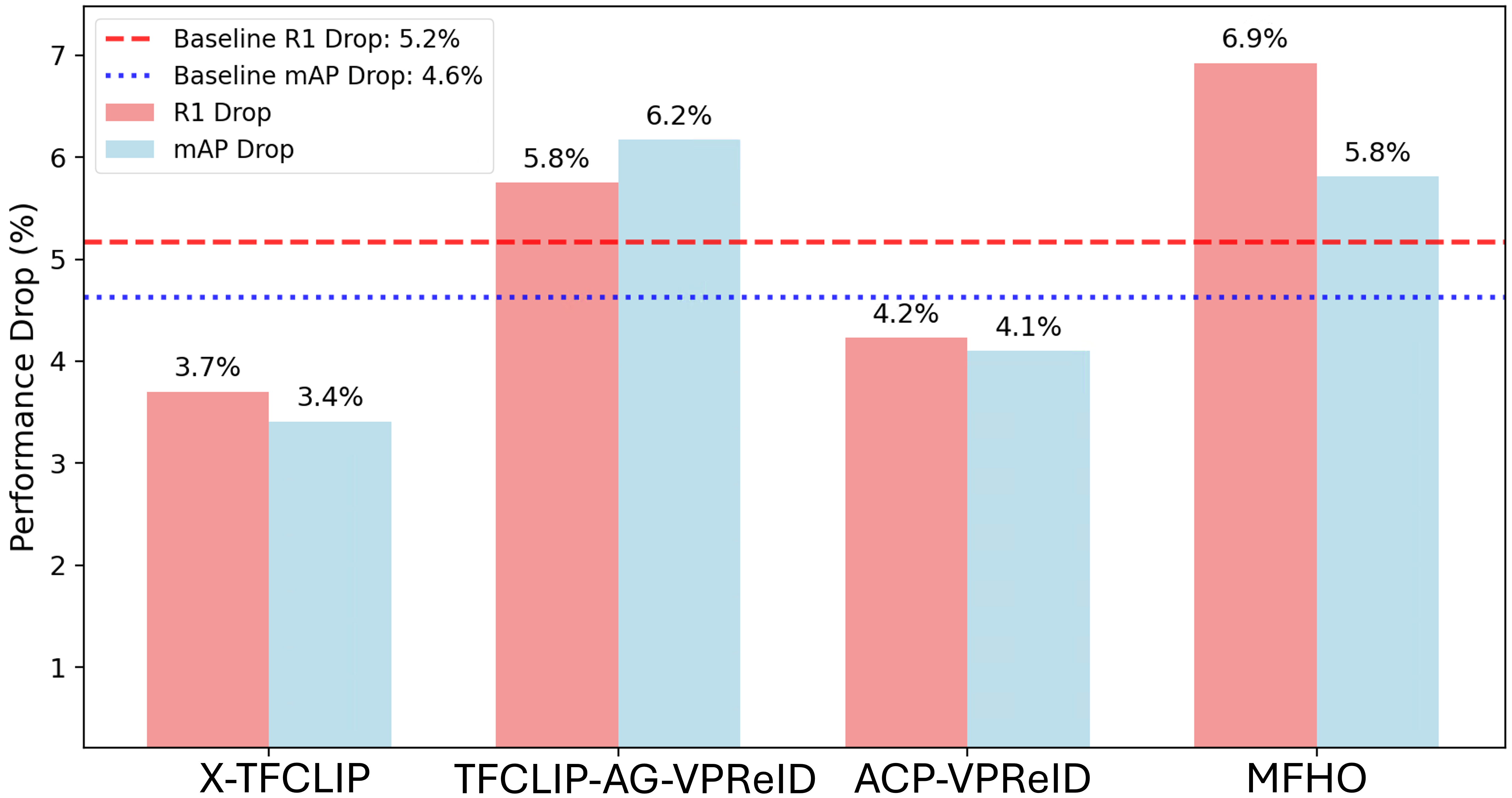}
    \caption{Performance drop 80m $\rightarrow$ 120m.}
    \label{fig:drop}
\end{figure}

\begin{figure}
    \centering
    \includegraphics[width=\linewidth]{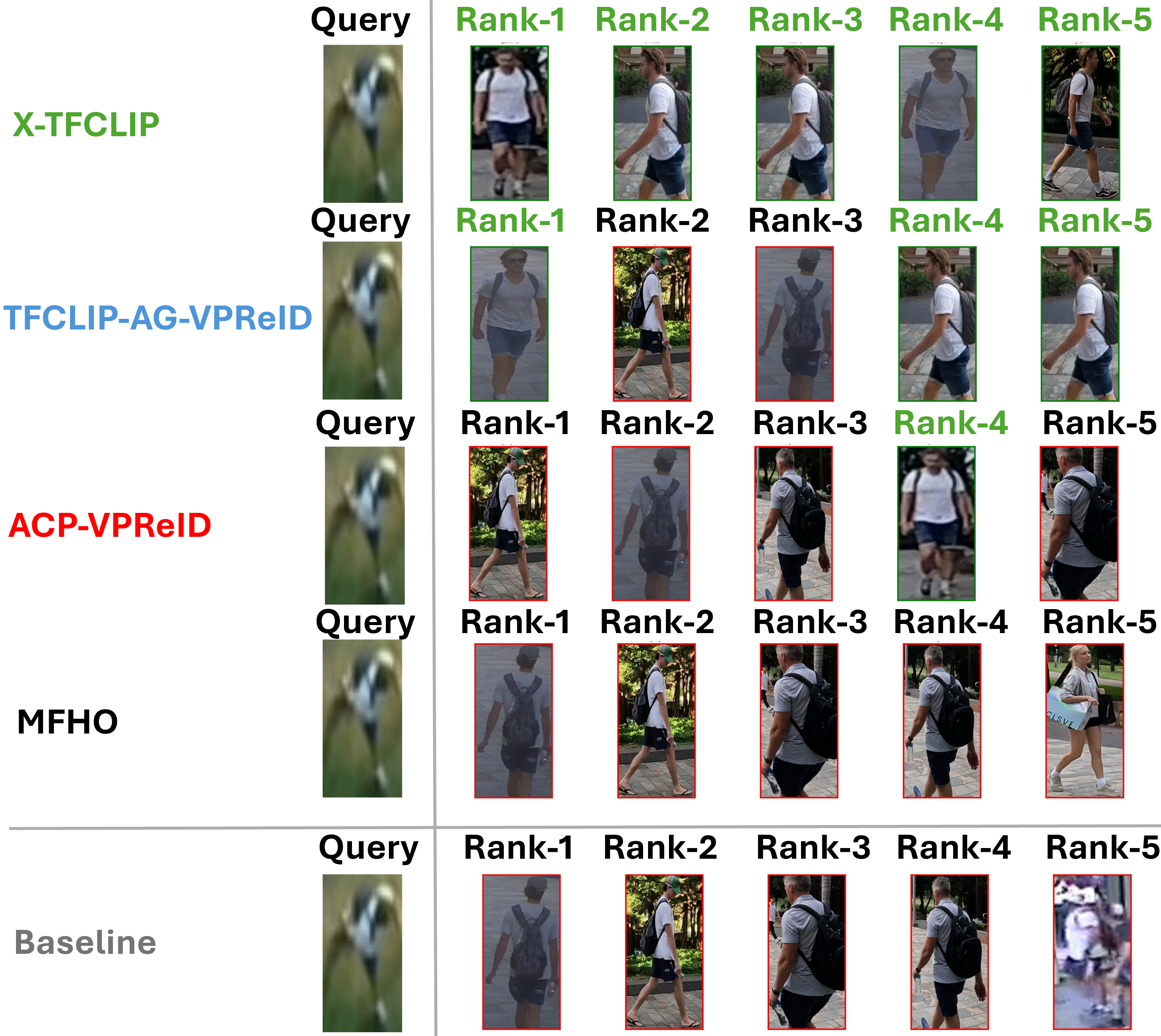}
    \caption{Aerial$\rightarrow$Ground test case at 120m altitude. Green/red: correct/incorrect labels.}
    \label{fig:vis_a2g_120}
\end{figure}

\textbf{Altitude-Specific Performance and Scale Challenges.} The comparison between 80m and 120m altitude performance reveals critical insights into the scalability of current approaches. TFCLIP-AG-VPReID excelled at 80m altitude with 78.47\% overall Rank-1, demonstrating the effectiveness of its stronger EVA-CLIP-L-14 backbone in handling moderate altitude scenarios, while X-TFCLIP maintained superior robustness at 120m with 73.58\% overall Rank-1 accuracy. Most methods experienced 4-6\% Rank-1 accuracy reduction when transitioning from 80m to 120m operations, as illustrated in Figure \ref{fig:drop}, highlighting severe challenges from extreme aerial viewpoints.

\textbf{Cross-Platform Matching and Temporal Dynamics.} An interesting observation is that Ground-to-Aerial matching scenarios generally outperformed Aerial-to-Ground scenarios across all methods, with TFCLIP-AG-VPReID achieving 73.28\% and 70.63\% Rank-1 accuracy respectively, showcasing the method's particular strength in upward viewpoint adaptation scenarios. This asymmetry suggests that upward viewpoint adaptation may be more tractable than downward perspective matching. The transition from image-based to video-based aerial-ground ReID introduces additional complexity through temporal dynamics, with leading methods demonstrating various approaches for handling these challenges.

\begin{figure}
    \centering
    \includegraphics[width=\linewidth]{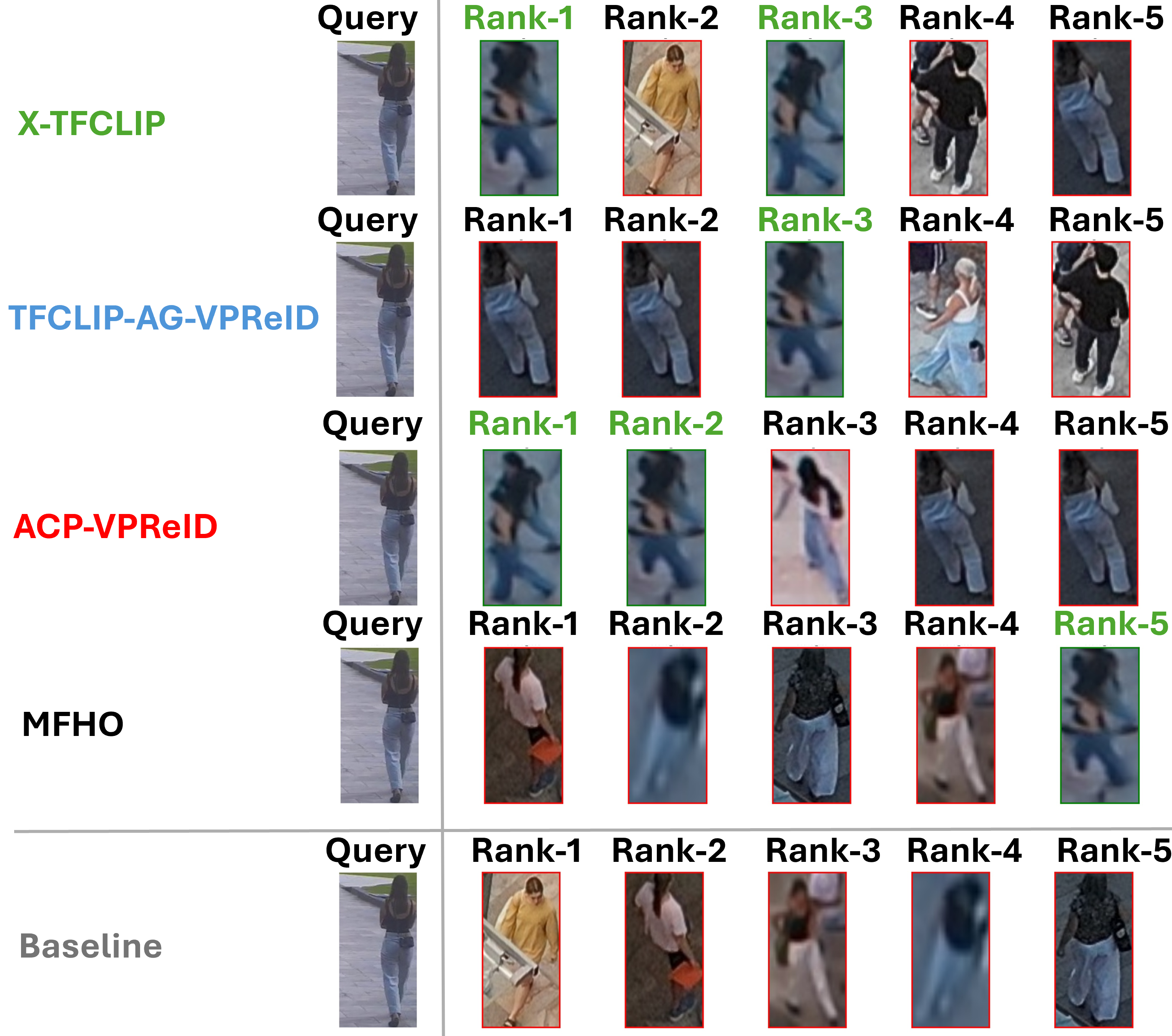}
    \caption{Ground$\rightarrow$Aerial test case at 120m altitude. Green/red: correct/incorrect labels.}
    \label{fig:vis_g2a_120}
\end{figure}

\textbf{Performance Gap and Research Opportunities.} While all four submitted methods demonstrated substantial improvements over the baseline TF-CLIP approach, the overall Rank-1 performance (64.93\%-71.89\%) remains significantly lower than those achieved in traditional ground-based video ReID scenarios (typically $>$90\%). This performance gap underscores the dataset's challenging nature and highlights substantial research opportunities in high-altitude aerial-ground person ReID, as shown in Figures~\ref{fig:vis_a2g_120} and~\ref{fig:vis_g2a_120}.

\section{Conclusions}
\label{sec:conclusion}
This paper presents the results of the AG-VPReID2025 Challenge, the first comprehensive evaluation for video-based person ReID across high-altitude aerial-ground scenarios. X-TFCLIP~(Appendix \ref{app:X-TFCLIP}) achieved top performance (71.56\% Rank-1, 73.60\% mAP) through attention-based feature selection, followed by TFCLIP-AG-VPReID~(Appendix \ref{app:DUT_IIAU_LAB}) (71.89\% Rank-1, 72.70\% mAP) which demonstrated competitive results through stronger EVA-CLIP-L-14 backbone architecture and expanded training dataset, ACP-VPReID~(Appendix \ref{app:ACP-VPReID}) (67.72\% Rank-1, 69.34\% mAP), and MFHO~(Appendix \ref{app:MFHO}) (64.93\% Rank-1, 67.32\% mAP). Despite technical advances, performance remains significantly lower than ground-based video ReID, highlighting challenges from extreme altitude (80–120m), resolution degradation, and scale variations. Altitude analysis revealed a 4-6\% performance drop from 80m to 120m, with Ground-to-Aerial matching outperforming Aerial-to-Ground scenarios, particularly evident in TFCLIP-AG-VPReID's superior G2A performance. This benchmark establishes a challenging evaluation platform driving research toward robust video-based person ReID under extreme operational conditions.

\appendix
\section{Submitted Person ReID Algorithms}\label{appendix}
In the following appendix, we provide a concise yet comprehensive summary of each video-based person ReID algorithm that was rigorously assessed during the AG-VPReID2025 Challenge.

\subsection{X-TFCLIP: An Extended TF-CLIP Method Adapted to Aerial-Ground Person Re-Identification }
\label{app:X-TFCLIP}
\noindent\emph{Tamás Endrei, Ivan DeAndres-Tame, Ruben Tolosana, Ruben Vera-Rodriguez, Aythami Morales, Julian Fierrez, Javier Ortega-Garcia} 
%\emph{endrei.tamas@itk.ppke.hu, ivan.deandres@uam.es, ruben.tolosana@uam.es, ruben.vera@uam.es, aythami.morales@uam.es, julian.fierrez@uam.es, javier.ortega@uam.es} \\

\begin{figure}[t]
\centering
\includegraphics[width=\linewidth]{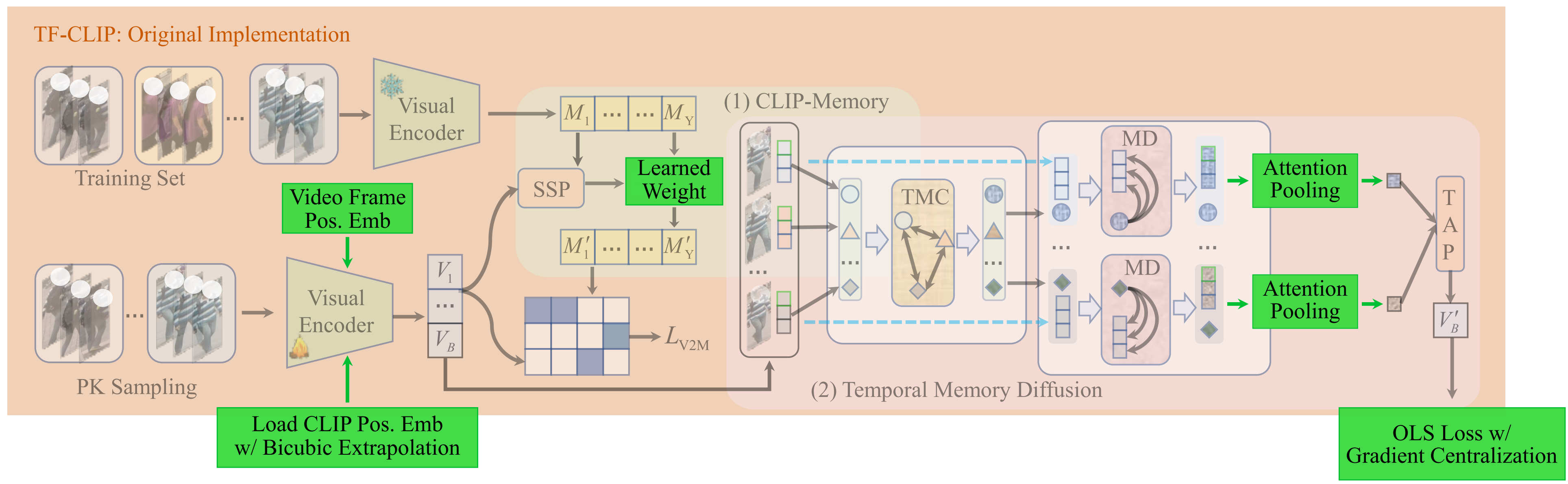}
\caption{Overview of our proposed X-TFCLIP architecture. The green modules highlight the key enhancements introduced over the original TF-CLIP~\cite{yu2024tfclip}.}
\label{fig:overview}
\end{figure}

\noindent{\textbf{Description of the algorithm:}} 
X-TFCLIP\footnote{\url{https://github.com/BiDAlab/X-TFCLIP}} is an extended version of TF-CLIP \cite{yu2024tfclip}, developed specifically for video-based person ReID in aerial-ground scenarios. Fig. \ref{fig:overview} shows a graphical representation of our proposed X-TFCLIP method, highlighting in green color the main improvements over the original TF-CLIP approach. The main contributions over TF-CLIP can be grouped into three key areas: \textit{i}) attention-based feature selection, \textit{ii}) architectural modifications to the CLIP backbone, and \textit{iii}) a revised loss function. In addition to these improvements, we also introduce adjustments at the inference stage and perform improved hyperparameter tuning to enhance training speed and stability.

First, in order to improve the temporal understanding of the model, X-TFCLIP replaces simple feature aggregation with an attention pooling mechanism that learns to select and weight the most informative frames in a sequence. This allows the model to focus on the most discriminative temporal cues when forming identity embeddings.

Regarding feature extraction, although our proposed X-TFCLIP method uses the same CLIP-ViT-B/16 backbone as the baseline, we introduce several enhancements to this encoder. First, we propose bicubic extrapolation for the model's absolute positional embeddings instead of bilinear, providing smoother spatial representations across varying video resolutions. Next, we change the normalization strategy in the BNNeck \cite{Luo2019BagOT} module from Batch Normalization \cite{Ioffe2015BatchNA} to Instance Normalization to handle appearance variations typical in person ReID. Additionally, we introduce a learnable frame-level positional embedding that is combined with camera embeddings to enrich the temporal and viewpoint context of each input frame. Finally, we replace the combination method between the CLIP Memory and projected visual features with a learnable weight parameter $\alpha$, allowing the model to adaptively control the contribution of each component.

In terms of training losses, X-TFCLIP employs Online Label Smoothing (OLS) \cite{zhang2021delving} in the CrossEntropy loss to generate soft target distributions, enhancing robustness and prediction calibration. To further improve feature discrimination, Triplet Loss and Image-to-Text (I2T) Loss are incorporated into the training pipeline. Finally, Gradient Centralization \cite{yong2020gradient} is applied as an optimization technique to promote faster and more stable convergence during training.

Regarding inference, we use the given soft-biometric labels to mask query-gallery pairs whose values are not the same, keeping unmasked the ones that were unknown. More precisely, we have used the gender, age, ethnicity, and body shape soft biometric labels for the masking. Moreover, we apply horizontal flipping to each tracklet to generate two embeddings, one from the original and one from the flipped version, and then average these embeddings to obtain the final representation for each tracklet.

\noindent \textbf{Experimental environment:}
The authors of this algorithm have confirmed that they adhered to the guidelines and rules specified in the \href{https://www.kaggle.com/competitions/agvpreid25}{AG-VPReID2025 challenge} during their evaluation. They also stated that they did not make any changes to the obtained results that would breach the challenge rules.

For the training of our X-TFCLIP method, only the provided AG-VPReID dataset is used, no further datasets or annotations. The CLIP backbone is the same as in the baseline, and thus is pretrained on ImageNet. The batches are created via PK sampling with $P = 4$ identities with $K = 6$ tracklets each, with a sequence length of 8 frames. The tracklet frames are resized to 288 $\times$ 144 pixel sizes. Regarding the soft-biometric masking, we set the mask values for the ethnicity, age, and gender to 0.2, and the mask value for the body shape to be 0.1. To match query and gallery tracklets, we compute the Euclidean distance between normalized embeddings and add the soft-biometric masking values to obtain the final distance matrix. All other training parameters not mentioned here are kept as in the original TF-CLIP method \cite{yu2024tfclip}. We trained the X-TFCLIP model for 90 epochs on an NVIDIA A-100 with 40GB of memory.

\subsection{Learning Text-Free CLIP for Aerial-Ground Video-based Person Re-identification with Stronger Backbone and Expanded Dataset (TFCLIP-AG-VPReID)}
\label{app:DUT_IIAU_LAB}
\noindent\emph{Zijing Gong, Yuhao Wang, Xuehu Liu, Pingping Zhang} 
%\emph{g2367941480@gmail.com, 924973292@mail.dlut.edu.cn, liuxuehu@whut.edu.cn, zhpp@dlut.edu.cn} \\

\begin{figure}[t]
\centering
\includegraphics[width=1\linewidth]{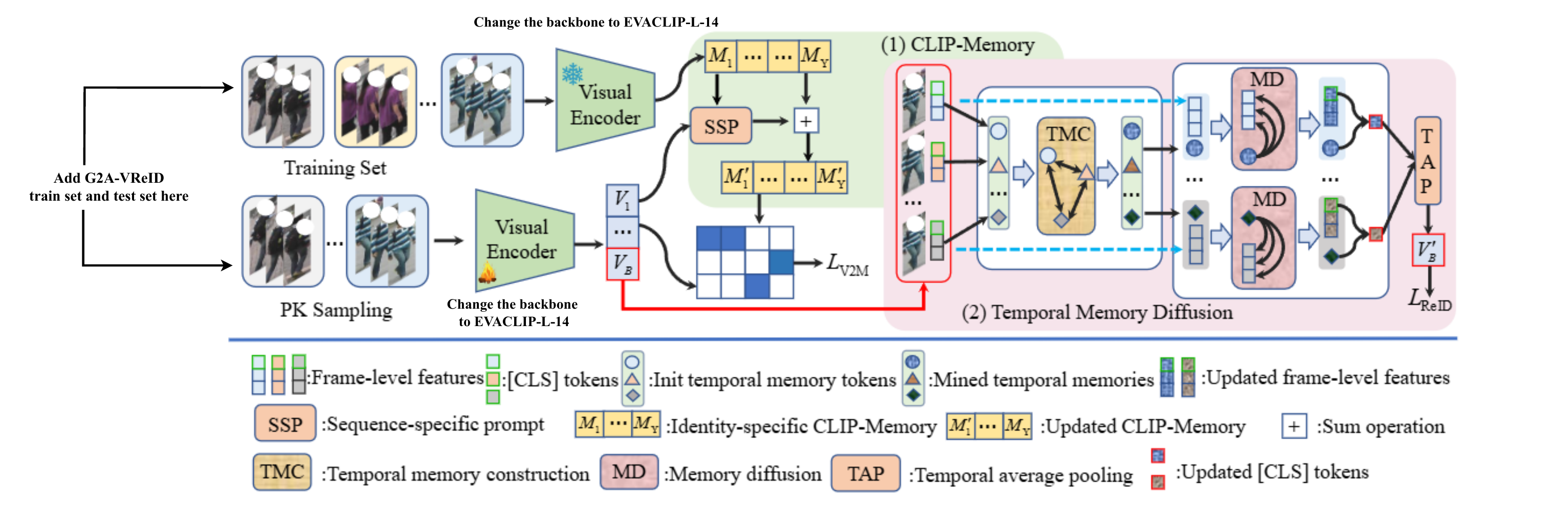}
\caption{Overview of the proposed TFCLIP-AG-VPReID. The model is built upon the TFCLIP baseline framework, replacing the original \texttt{ViT-B-16} image encoder with an \texttt{EVA-CLIP-L-14} encoder. The training dataset is expanded by combining the AG-VPReID 2025 training set with the G2AVReID dataset.}
\label{fig:over}
\end{figure}

\noindent \textbf{Description of the algorithm:} As shown in Fig.~\ref{fig:over}, our algorithm is derived from the TFCLIP baseline framework. We replace the original \texttt{ViT-B-16} image encoder with an \texttt{EVA-CLIP-L-14} encoder while keeping all other modules unchanged. 
This change allows the model to extract richer semantic features from input images. 
To meet the increased data requirements of the new backbone network, we expand the training dataset by incorporating both the training and test sets of the G2AVReID dataset into the original AG-VPReID 2025 dataset. 
Person identities are reorganized such that new IDs continue sequentially after the maximum ID value from the original dataset—for example, if the highest ID in the original dataset is 689, the new IDs begin at 690, 691, and so on.
The total loss function is defined as the sum of three component losses:
\begin{equation}
    \mathcal{L}_{\text{total}} = \mathcal{L}_{V2M} + \mathcal{L}_{tri} + \mathcal{L}_{ce},
\end{equation}
where $ \mathcal{L}_{V2M} $ denotes the Video-to-Memory Contrastive Loss, $ \mathcal{L}_{tri} $ represents the Triplet Loss, and $ \mathcal{L}_{ce} $ corresponds to the Label Smoothing Cross-Entropy Loss. Each loss function contributes equally to the final objective with a weight of 1.0.
The Video-to-Memory Contrastive Loss encourages the model to align video tracklets with their corresponding memory representations and is defined as:
\begin{equation}
    \mathcal{L}_{V2M}(y_i) = -\frac{1}{|P(y_i)|} \sum_{p \in P(y_i)} \log \frac{\exp(\text{Cos}(\mathbf{v}_p, \mathbf{M}'_{y_i}))}{\sum_{j=1}^{B} \exp(\text{Cos}(\mathbf{v}_j, \mathbf{M}'_{y_i}))},
\end{equation}
where $ \mathbf{v}_p $ represents the feature vector of a positive sample, $ \mathbf{M}'_{y_i} $ is the memory representation for identity $ y_i $, and $ B $ is the batch size.
The Triplet Loss enhances feature discriminability by increasing the distance between embeddings of different identities and reducing it for the same identity. It is defined as:
\begin{equation}
    \mathcal{L}_{tri} = \max(\|\mathbf{f}_a - \mathbf{f}_p\|^2_2 - \|\mathbf{f}_a - \mathbf{f}_n\|^2_2 + \alpha, 0),
\end{equation}
where $ \mathbf{f}_a $ is the anchor embedding, $ \mathbf{f}_p $ is the positive embedding, $ \mathbf{f}_n $ is the negative embedding, and $ \alpha $ is the margin parameter set to $ 0.3 $.
The Label Smoothing Cross-Entropy Loss improves generalization by softening the one-hot label distribution. It is defined as:
\begin{equation}
    \mathcal{L}_{ce} = -\sum_{k=1}^{Y} q(k) \log(p(k)),
\end{equation}
where $ q(k) $ is the smoothed label distribution, and $ p(k) $ is the predicted probability. The smoothing factor $ \epsilon $ is set to $ 0.1 $.
Additionally, batch hard mining is applied during training to focus on difficult samples within each mini-batch.
With stronger backbone features and an expanded training dataset, our model achieves a significant improvement in performance over the baseline in the AG-VPReID 2025 challenge, demonstrating the effectiveness of these modifications.

\noindent\textbf{Experimental environment:} 
For implementation, we build our pipeline using PyTorch 2.0.0. 
We use the AG-VPReID 2025 dataset and the G2AVReID dataset for training, and the AG-VPReID 2025 dataset for testing. 
The model is trained on an NVIDIA A800 GPU for a total of 80 epochs, taking approximately 40 hours to complete. 
The batch size is set to 8, containing 8 frames per track, sampled from 2 identities, each with 4 tracks. 
Input images are resized to $ 224 \times 224 $ pixels. 
The learning rate schedule includes a warm-up phase over the first 10 epochs, where it linearly increases from $ 5 \times 10^{-7} $ to $ 5 \times 10^{-6} $. 
Subsequently, the learning rate is divided by 10 at the 30th and 50th epochs to ensure stable convergence.

\subsection{Adaptive Cross-Perspective Video Person Re-Identification Network (ACP-VPReID)}
\label{app:ACP-VPReID}
\noindent\emph{Md Rashidunnabi, Hugo Proença, Kailash A. Hambarde} 
%\emph{md.rashidunnabi@ubi.pt, hugomcp@ubi.pt, n1947@ubi.pt} \\

\noindent \textbf{Description of the algorithm:} 
Our method, {ACP-VPReID}, extends the TF-CLIP~\cite{Yu2023TF} baseline to better handle cross-view video ReID between aerial and ground platforms. Key innovations include \textit{(i) View-specific feature encoding:} Tracklets from different views (aerial, ground, wearable) are processed through dedicated transformation layers to preserve viewpoint-specific semantics. \textit{(ii) Temporal Memory Diffusion:} We introduce view-aware memory banks that store long-term temporal features and use gated attention to adaptively retrieve relevant information. \textit{(iii) Motion-aware gating:} Temporal differences between frames are used to emphasize dynamic content and suppress static noise. \textit{(iv) Scale-adaptive modules:} To handle size discrepancies across views, we apply scale normalization layers conditioned on camera type. \textit{(v) Enhanced prompting:} Learnable view embeddings and cross-view transformers guide feature aggregation across time and perspective. These components work together to produce robust, disentangled, and view-adaptive representations that significantly improve retrieval accuracy under extreme viewpoint shifts.

\noindent \textbf{Experimental environment:}
We followed the official \href{https://www.kaggle.com/competitions/agvpreid25}{AG-VPReID2025 challenge} rules, using only the provided training set~\cite{nguyen2025ag} without external data. Our backbone is CLIP ViT-B/16~\cite{radford2021learning}, extended with the TF-CLIP framework~\cite{Yu2023TF} and our proposed view-adaptive modules.
Experiments were run on a single NVIDIA A40 GPU using PyTorch 1.11.0 with CUDA 12.4 and automatic mixed precision. We used deterministic training with fixed random seeds.

Training setup includes the following configurations: \textit{(i) Input:} 8-frame clips, resized to $256\times128$, normalized, with random flip, erase, and padding. \textit{(ii) Optimization:} AdamW, initial LR $1\times10^{-4}$ (stage 1), reduced to $5\times10^{-6}$ (stage 2), batch sizes 64/32, dropout 0.3, weight decay 0.05. \textit{(iii) Losses:} identity, triplet, center, and contrastive loss (image-text via CLIP clustering). \textit{(iv) Augmentation:} random flip (80\%), erase (60\%), rectangular scaling, ImageNet normalization.

Model configuration parameters are as follows: \textit{(i) Patch size:} $16\times16$, feature dim: 768, projection: 512. \textit{(ii) View-specific modules:} memory banks, classification heads, and visual prompts. \textit{(iii) Attention:} 12 heads, view-aware multi-head self-attention and cross-view transformers. \textit{(iv) Training:} 150 epochs (stage 1), 100 epochs (stage 2) with evaluation every 5/2 epochs respectively.

\subsection{Multi-Fidelity Hyperparameter Optimization (MFHO)}
\label{app:MFHO}
\noindent\emph{Saeid Rezaei} 
%\emph{Saeid.Rezaei@ucc.ie} \\

\noindent {\textbf{Description of the algorithm:}} Multi-fidelity optimization is a strategy in hyperparameter optimization that uses lower-cost evaluations (such as training models with fewer epochs or using subsets of data) to rapidly narrow down hyperparameter search spaces, subsequently applying higher-cost evaluations to promising candidates. This approach significantly reduces computational resources compared to exhaustive evaluations \cite{falkner2018bohb}. Multi-fidelity methods are particularly advantageous in machine learning, especially deep learning, due to the high computational cost associated with training models \cite{li2018hyperband}. The Python code provided employs a simplified variant of the Hyperband algorithm. Initially, random configurations are tested with limited fidelity (fewer epochs), rapidly discarding poor configurations. Promising configurations identified from these lower-fidelity runs are tested more thoroughly to confirm their performance. The algorithm follows these steps: random sampling to select hyperparameter configurations, lower-fidelity evaluation to quickly assess model performance, performance assessment to compare with the current best, and higher-fidelity evaluation to reassess promising configurations when needed. This iterative, fidelity-aware process allows rapid and resource-efficient exploration of the hyperparameter space.

\noindent\textbf{Experimental environment:} The authors of this algorithm confirmed their adherence to the guidelines and rules specified in the \href{https://www.kaggle.com/competitions/agvpreid25}{AG-VPReID2025 challenge} during evaluation, stating that no modifications were made to the obtained results that would breach the regulations. The experimental setup utilized the AG-VPReID2025 training set with CUDA version 12.2 and the Python programming language.

% \begin{figure*}
%     \centering
%     \includegraphics[width=0.9\linewidth]{figs/a2g_80m.png}
%     \caption{Aerial to Ground 80m}
%     \label{fig:vis_a2g_80}
% \end{figure*}

% \begin{figure*}
%     \centering
%     \includegraphics[width=0.9\linewidth]{figs/g2a_80m.png}
%     \caption{Ground to Aerial 80m}
%     \label{fig:vis_g2a_80}
% \end{figure*}

\section{Acknowledgements}
This work was supported by QUT's Research Engineering Facility, ARC Discovery Grants DP200101942 and DP250103634, and QUT Postgraduate Research Award. The X-TFCLIP method development was supported by INTER-ACTION (PID2021-126521OBI00 MICINN/FEDER), C\'{a}tedra ENIA UAM-VERIDAS en IA Responsable (NextGenerationEU PRTR TSI-100927-2023-2), and PowerAI+ (SI4/PJI/2024-00062, Comunidad de Madrid and UAM), and R\&D Agreement DGGC/UAM/FUAM for Biometrics and Cybersecurity. We thank all participating teams, drone operators, and volunteers for their contributions.

%\section{To be Sorted}

{\small
\bibliographystyle{ieee}
\bibliography{egbib}
}

\end{document}